\documentclass[conference]{IEEEtran}
\IEEEoverridecommandlockouts
\usepackage{cite}
\usepackage{ragged2e}

\usepackage{amsmath,amssymb,amsfonts}
\usepackage{algorithmic}
\usepackage{graphicx}
\usepackage{textcomp}
\usepackage{booktabs}
\usepackage{xcolor}
\usepackage{hyperref}
\def\BibTeX{{\rm B\kern-.05em{\sc i\kern-.025em b}\kern-.08em
    T\kern-.1667em\lower.7ex\hbox{E}\kern-.125emX}}
\begin{document}

\title{Connecting the Dots: Graph Neural Network Powered Ensemble and  Classification of Medical Images\\
\thanks{Supported by Science Foundation Ireland under Grant No. 18/CRT/6049}
}

\author{\IEEEauthorblockN{1\textsuperscript{st} Aryan Singh}
\IEEEauthorblockA{\textit{Dept. of Electronic and Computer Engineering} \\
\textit{University of Limerick}\\
Limerick, Ireland \\
aryan.singh@ul.ie}
\\
\IEEEauthorblockN{3\textsuperscript{rd} Ciarán Eising}
\IEEEauthorblockA{\textit{Dept. of Electronic and Computer Engineering} \\
\textit{University of Limerick}\\
Limerick, Country \\
ciaran.eising@ul.ie}
\and
\IEEEauthorblockN{2\textsuperscript{nd} Pepijn Van de Ven}
\IEEEauthorblockA{\textit{Dept. of Electronic and Computer Engineering}\\
\textit{University of Limerick}\\
Limerick, Ireland \\
pepijn.vandeven@ul.ie}
\\
\IEEEauthorblockN{4\textsuperscript{th} Patrick Denny}
\IEEEauthorblockA{\textit{Dept. of Electronic and Computer Engineering} \\
\textit{University of Limerick}\\
Limerick, Ireland \\
patrick.denny@ul.ie}
}

\IEEEoverridecommandlockouts
\IEEEpubid{\makebox[\columnwidth]{ 979-8-3503-6021-9/23/\$31.00~\copyright2023 IEEE \hfill} \hspace{\columnsep}\makebox[\columnwidth]{ }}

\maketitle

\IEEEpubidadjcol

\begin{abstract}
Deep learning models have demonstrated remarkable results for various computer vision tasks, including the realm of medical imaging. However, their application in the medical domain is limited due to the requirement for large amounts of training data, which can be both challenging and expensive to obtain. To mitigate this, pre-trained models have been fine-tuned on domain-specific data, but such an approach can suffer from inductive biases. Furthermore, deep learning models struggle to learn the relationship between spatially distant features and their importance, as convolution operations treat all pixels equally. Pioneering a novel solution to this challenge, we employ the Image Foresting Transform to optimally segment images into superpixels. These superpixels are subsequently transformed into graph-structured data, enabling the proficient extraction of features and modeling of relationships using Graph Neural Networks (GNNs). Our method harnesses an ensemble of three distinct GNN architectures to boost its robustness. In our evaluations targeting pneumonia classification, our methodology surpassed prevailing Deep Neural Networks (DNNs) in performance, all while drastically cutting down on the parameter count. This not only trims down the expenses tied to data but also accelerates training and minimizes bias. Consequently, our proposition offers a sturdy, economically viable, and scalable strategy for medical image classification, significantly diminishing dependency on extensive training data sets. Our code is available at \href{https://github.com/aryan-at-ul/AICS_2023_submission}{Github}.
\end{abstract}


\section{Introduction}
Deep Neural Networks (DNNs) have been proven to be effective for computer vision tasks and are increasingly being utilized in medical imaging research. These models have evinced state-of-the-art performance in an array of tasks including object detection, image classification \cite{726791}, semantic segmentation, and instance segmentation \cite{Sultana_2020}. Acquiring labeled data in the medical domain can be an arduous and costly undertaking, and data quality may vary considerably across different sources. Consequently, an ideal model would require less training data and fewer parameters, leading to greater efficiency and reduced computational resources. Moreover, such models would offer superior generalizability \cite{ba2014deep}, a crucial feature that is often limited in DNNs. While DNNs can perform well on the data they were trained on, they may struggle to generalize to new and unseen data, which highlights a critical issue with the use of DNNs in the medical domain \cite{Zhou_2021} where labeled data of high quality can be scarce.

Notably, while DNNs employ convolutional kernels for fixed local pixel grid connectivity and pooling for global feature extraction, they overlook the inherent topological structure of medical images. This omission, crucial for optimal medical image understanding and representation, has been highlighted \cite{poscnn,spatcnn}.

Graph-based neural networks (GNNs) adeptly handle variable-sized heterogeneous graph input \cite{8961143}, allowing for adaptability across diverse data and tasks. These models can discern intricate geometric interrelationships in image datasets \cite{Bronstein_2017}, enhancing predictive performance \cite{9693311}. GNNs, capable of analyzing complex interconnected phenomena, are increasingly used in medical imaging \cite{shen2022gnns}. Their application in medical imaging, including classification \cite{Ahmedt_Aristizabal_2021} and segmentation \cite{histgnn}, has seen significant progress. GNNs have proven effective in multi-modal data-based medical image analysis \cite{ding2022graph,keicher2021ugat} and in human-object interaction detection using deep neural network (DNN) features \cite{liang2021visualsemantic}. Motivated by these advancements, we aim to explore the synergistic use of DNN features and GNNs in transducing medical images into graphs.

In this study, GNNs were applied to classify pneumonia from medical images of the lungs and have demonstrated effectiveness in this task. Our approach utilizes a method that involves transforming X-ray images into graphs, taking into account the topological connections between the various features present. This graph structure not only enhances the global level representation captured by DNNs, but also provides a more comprehensive understanding of the image's overall structure. By leveraging this graph structure, our approach can effectively capture both locally and globally relevant features, resulting in improved performance in X-ray image analysis tasks. Our study presents a compelling comparison between the performance of GNNs and three prominent DNNs, which serves to highlight the significant advantages of incorporating GNNs alongside our method for graph creation. By showcasing the improved performance achieved through the use of GNNs, we provide strong evidence for the effectiveness of our approach and its potential for application in the medical imaging domain. We made efforts to enhance performance by ensembling multiple GNN architectures trained on varying graph sizes, reaching a sensitivity score as high as 99\%, a critical factor in the medical domain. By ensembling different GNNs, we can leverage the strengths of each model while mitigating their individual weaknesses, ultimately yielding a more robust and accurate system.




\section{Methods}
Our method is divided into two distinct stages.

\begin{figure*}[tb]
 \centering
 \includegraphics[width=1\textwidth,height=1.75in]{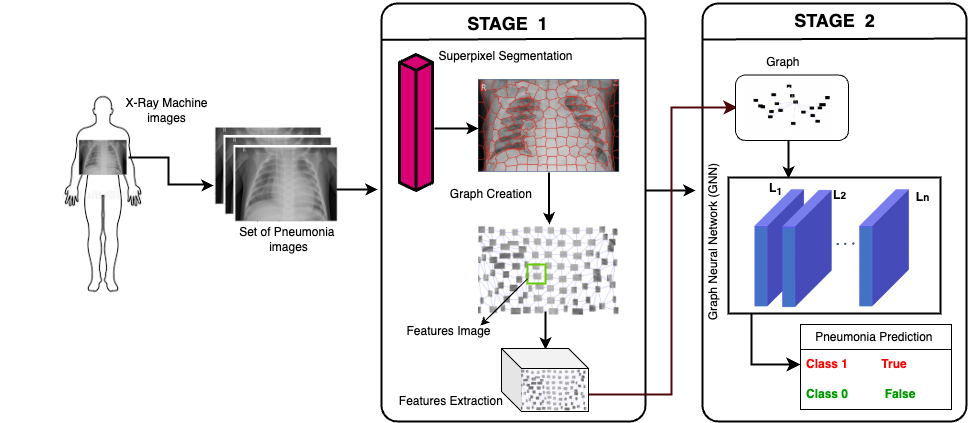}
 \caption{Image to graph pipeline stage 1 and stage 2}
 \label{fig:stageb}
\end{figure*}

The first stage, as illustrated in Figure \ref{fig:stageb}, involves dividing the input image into smaller, homogeneous regions called superpixels by grouping adjacent pixels based on aggregating criteria such as color, intensity, or texture similarity. We then create a graph with these superpixels as nodes and establish edges between nodes based on region adjacency and homogeneity explained in section \ref{sec:gc}. The features for each superpixel are obtained using a DNN. The extracted features from the superpixel segments of the image. including basic features like edges, corners, and textures from the early stages of the network. As the image gets passed through the deeper layers of the network, it extracts more complex and abstract features such as objects, scenes, and different parts of objects. These features are then used as node features in the graph.

The second stage, as depicted in Figure \ref{fig:stageb}, classifies the entire graph for pneumonia. Utilizing a GNN for graph classification generates distinct embeddings for pneumonia and non-pneumonia graphs, which incorporate the features extracted using a DNN as node features. The aggregation of neighbouring node features contributes to better embedding, enhancing graph-level classification performance.




\subsection{Superpixel Segmentation}
We have used superpixel segmentation to divide an image into smaller regions, that are more homogeneous in terms of colour, texture, and other features. Superpixel segmentation overcomes the limitations of primitive segmentation methods, which tend to produce irregular and fragmented regions. Using superpixel segmentation, we effectively isolate regions of interest and obtain a more detailed understanding of the underlying image structure. To generate superpixels, we have implemented two algorithms: Simple Linear Iterative Clustering (SLIC) \cite{slic} and Dynamic and Iterative Spanning Forest (DISF) \cite{Belem_2020}.

\subsubsection{SLIC}
The SLIC algorithm begins with dividing the image into a grid of small, rectangular cells of size $S \times S$ where $S$ = $\sqrt{\frac{N}{k}}$ where $N$ is the number of pixels and $k$ is the desired number of superpixels. The input to the SLIC algorithm is the number $k$. Each pixel is then assigned to one of the initial clusters which are sampled on a regular grid spaced $S$ pixels apart. In the case of images in the CIELAB colour space, each pixel is allocated to a cluster based on its distance to the cluster centre, represented by the tuple $C_{i} = [l_{i}, a_{i}, b_{i}, x_{i}, y_{i}]$, where $l_{i}$ is the lightness, $a_{i}$ is the colour axis ranging from green to red, $b_{i}$ is the colour axis ranging from blue to yellow, and $x_{i}$, $y_{i}$ are the cluster centre coordinates. Each pixel is associated with its nearest cluster. Once all pixels have been assigned a cluster,  the cluster centre is adjusted to the mean of the $[l,a,b,x,y]^T$ vector applied to all pixels belonging to the cluster. One notable advantage of SLIC over other clustering methods is that it only considers pixels in a small, local region (of size $2S \times 2S$) when forming clusters, which makes the algorithm faster to run. 

\subsubsection{DISF}
The DISF algorithm takes two inputs: initial seed value $s$ and the number of superpixels $k$. An image (I) is defined as a set of pixels $D_{I}$, and a function $I(p)$ that assigns local attributes to each pixel $p$ in $D_{I}$. Seeds are placed at equal distances in a grid format, with distance $d$ = $\sqrt{\frac{N}{k}}$, where $N$ is a set of the neighbourhood nodes of pixel $p_{i}$.
$N \subseteq  D_{I}$ and $k$ is the desired number of superpixels. Thus one can define an image as a graph $\mathcal{G}$ as $(N,A,I)$ where $A$ is the adjacency relation $A \subset N 
 \times N$.

DISF leverages the Image Foresting Transform (IFT)\cite{ift} to achieve better superpixel segmentation.

At the core of the DISF algorithm, the IFT method involves the iterative construction of minimum spanning trees using the image as a graph with seeds as nodes. In each iteration, the superpixel assignments are refined by adjusting the pixel similarity measures. This is accomplished by minimizing the cost map across all paths in the tree. The iterations continue, progressively enhancing superpixel assignments until convergence is reached or a predefined maximum number of iterations is completed.

One of the most salient benefits of the DISF algorithm lies in its ability to surmount the limitations of conventional superpixel methods, particularly in terms of initial seed availability. By employing an oversampling strategy, DISF increases the initial seed count, improving the likelihood of including all relevant seeds and resulting in more representative superpixels. This, in turn, enhances the overall segmentation quality \cite{Vargas_Munoz_2019}. Additionally, DISF removes seeds that produce the smallest trees in homogeneous regions of the image, based on vectors representing each superpixel's properties and its neighboring superpixels. This ensures that the ultimate superpixels are more homogeneous and accurately represent the image's structure, culminating in superior segmentation outcomes.

\subsection{Graph Creation}
\label{sec:gc}

After segmenting an image into superpixels, we proceed to construct a Region Adjacency Graph (RAG). In this graph, each superpixel is represented as a node, and the nodes are connected to their neighboring nodes based on their spatial proximity within the image. These connections between the nodes are called edges.

To assign weights to these edges, we calculate the mean colour value of the pixels within each superpixel and then determine the differences in these mean values between adjacent superpixels. In order to augment the graph with additional information, we can incorporate features extracted from a DNN as node attributes \ref{sec:sfe}. These features can help to improve the graph's utility in subsequent image classification tasks.

Once the RAG has been generated and enhanced with the DNN extracted features, it is then saved as a pickle file, for easy retrieval and use in future classification tasks.
 
In the next sub-section, we will delve into the methods used for feature extraction.

\subsection{Superpixel Feature Extraction}
\label{sec:sfe}
Once we have generated the superpixels of an image, we use pre-trained DNNs for feature extraction. We have chosen ResNet18, DenseNet121, and EfficientNetB0 to extract features from the superpixel segments. To achieve this, we first generate a new image by merging the segmented mask region with a bounding rectangle, which is constructed using the minimum and maximum pixel coordinates of the respective segment. We then resize the derived image to a standardized dimension, ensuring compatibility with the input requirements of the DNNs during the feature extraction phase. 

We provide a detailed description of the employed DNNs for feature extraction below:

\subsubsection{ResNet18}
ResNet \cite{resnet} Deep Convolutional Neural Networks make use of stacked layers (depth) to capture low to high-level features. The depth of the network is important because it produces good results, but also introduces several challenges. One main problem that arises from stacking multiple layers is that of vanishing/exploding gradients \cite{vanexp}, which is partially solved by normalized initialization. However, as the network gets deeper, the accuracy starts to decline and training loss starts to increase. These problems arise primarily because, of the identity mapping of the features being propagated from one layer to another. It becomes difficult to optimize performance as the model becomes deeper. ResNet adds a skip connection that uses an identity function to bypass the non-linear activation. The advantage of this architecture is the flow of gradient through the identity function from the later layer to earlier layers, but the summation operation between the identity function and output from a layer impedes the information flow in the network. ResNet mitigates this issue by using residue blocks in DNNs, which do not add additional parameters to the model but achieve better accuracy. ResNets have been successfully applied in the medical imaging domain and have produced state-of-the-art results. For example, ResNet has been applied for medical image retrieval \cite{ayyachamy2019medical} and to breast cancer classification from histopathological images \cite{breast}.

\subsubsection{DenseNet121}
DenseNet, is a DNN architecture characterized by its unique connectivity pattern, which establishes connections between each layer and all other layers in a feed-forward manner. This dense connectivity ensures that the feature map of every preceding layer serves as input for the current layer, while its own feature map contributes as input to all subsequent layers.
The distinctive structure of DenseNet enables each layer to receive direct inputs from all preceding layers. These inputs undergo concatenation, followed by a series of three consecutive operations. The process begins with batch normalization (BN), which is designed to improve the stability and convergence of the network. Next, the rectified linear unit (ReLU) is applied as an activation function to introduce non-linearity and enhance the model's capacity to learn complex patterns. Finally, a 3 × 3 convolution (Conv) operation is performed to extract features from the input data, as described in the DenseNet literature \cite{densenet}. This combination of operations within DenseNet allows for efficient feature extraction and contributes to the overall effectiveness of our proposed framework.

\subsubsection{EfficientNetB0}
EfficientNet \cite{effnet} is an innovative approach to scaling deep neural network architectures, tailored to specific use cases depending on data or resource constraints. The key advantage of EfficientNet lies in its ability to enhance model performance without increasing the number of floating-point operations per second (FLOPS), making it a more efficient choice compared to other DNNs.

The primary concept behind EfficientNet is the introduction of a method to uniformly scale all dimensions of a neural network, including depth, width, and resolution, by employing efficient compound coefficients. By scaling a model's computational resources by a factor of $2^{N}$, EfficientNet adjusts the depth of the network by $\alpha^{N}$, the width of the network by $\beta^{N}$, and the image size by $\gamma^{N}$. These coefficients, $\alpha$, $\beta$, and $\gamma$, are determined through a grid search performed on the base model.

The compound scaling method employed by EfficientNet facilitates a well-balanced trade-off between the accuracy and efficiency of the networks, enabling users to optimize their models based on available resources. This ingenious approach is tailored such that the network architecture is adapted effectively to meet the specific needs of various use cases while maintaining computational efficiency.

\subsection{Graph Neural Networks}
For this study, we examined various types of spectral and spatial GNNs, including Graph Convolutional Neural Networks (GCNN), Graph Attention Networks, and Graph Isomorphism Networks. To address the limitations of individual models, we also tested combining all of these GNNs through ensemble methods. We provide a brief overview of each GNN in this section.

\subsubsection{Graph Convolutional Neural Networks (GCNN)}
GCNNs are a class of deep-learning models specifically designed for graph-structured data. They extend the concept of convolutional operations from the realm of grid-like data, such as images, to irregular domains represented by graphs. GCNNs can be broadly categorized into two approaches: spectral and spatial.

The spectral approach leverages the principles of graph signal processing and transforms the graph to the spectral domain by utilizing eigenvectors of the graph's Laplacian matrix. To simplfy filters, a learned a learnable diagonal matrix, $g_{w}$ is used, and various designs for the filter $g_{w}$ have been proposed. GCNNs employ techniques to avoid computing Laplacian eigenvectors for efficiency \cite{GCNN}.

In contrast, the spatial approach directly defines convolution on the graph based on its topology, using feature aggregation through message passing. The convolution is defined according to the number of hops allowed between a node and its neighbours \cite{sage}, aiming to maintain local invariance similar to traditional CNNs, despite varying neighbour sizes.

\subsubsection{Graph Attention Networks (GAT)}
The issue of over-smoothing has long plagued traditional GCNNs as the addition of multiple layers leads to simple graph embeddings for different class graphs. GAT addresses this problem by utilizing an attention mechanism to minimize over-smoothing. 

GAT \cite{gat} makes use of attention-based convolution operation, which assigns different weights for neighbours to make the learning process more robust and stable, thus alleviating the noise effects. The attention-based mechanism allows for dealing with variable-sized input and making use of the most relevant part of the input to make inferences.  GAT makes use of a self-attention layer and multi-head attention mechanisms \cite{attention}. 

The input to the attention layer is  \(\mathbf{h}=\left\{\vec{h}_{1}, \vec{h}_{2}, \ldots, \vec{h}_{N}\right\}, \vec{h}_{i} \in \mathbb{R}^{F}\), where \(N\) is the number of nodes, and \(F\) is the number of features in each node which produces the output \(\mathbf{h}^{\prime}=\left\{\vec{h}_{1}^{\prime}, \vec{h}_{2}^{\prime}, \ldots, \vec{h}_{N}^{\prime}\right\}, \vec{h}_{i}^{\prime} \in \mathbb{R}^{F^{\prime}}\). The self-attention mechanism is applied to every node, where the attention coefficient is computed as 

\[
e_{i j}=a\left(\mathbf{W} \vec{h}_{i}, \mathbf{W} \vec{h}_{j}\right)
\]

which absorbs the importance of node \(j\) 's features to node \(i\), this operation is executed across all nodes, it results in the loss of the inherent graph structure. This issue is overcome by performing masked attention over nodes,  \(e_{i j}\) for nodes \(j \in \mathcal{N}_{i}\), where \(\mathcal{N}_{i}\) is some neighbourhood of node \(i\) in the graph. For reason that first-order nodes
$N(v) = \{ u \in V \mid (u,v) \in E \}$ of a node can be of different sizes, and the coefficient is normalized across all choices of \(j\) using the softmax function

\[
\alpha_{i j}=\operatorname{softmax}_{j}\left(e_{i j}\right)=\frac{\exp \left(e_{i j}\right)}{\sum_{k \in \mathcal{N}_{i}} \exp \left(e_{i k}\right)} .
\]

The normalized attention coefficients are used to compute the linear combination of features of all nodes. The aggregated features from each head are concatenated or averaged to obtain \(\vec{h}_{1}^{\prime}\), and finally a non-linear activation is applied:

\[
\vec{h}_{i}^{\prime}=\sigma\left(\sum_{j \in \mathcal{N}_{i}} \alpha_{i j} \mathbf{W} \vec{h}_{j}\right) .
\]

\subsubsection{GRAPH ISOMORPHISM NETWORK (GIN)}
GIN \cite{xu2019powerful} was developed to test the power of GNNs, in comparison to the Weisfeiler-Lehman (WL) test of graph isomorphism \cite{wlm} which is an algorithm used to determine if two graphs are isomorphic, meaning they have the same structure and patterns, but with different labels. The WL algorithm uses a recursive procedure to generate hashes which are used to find the structural similarity between two graphs. GIN makes use of the WL algorithm in continuous feature space. GIN aggregates node features using a Learnable Neighborhood Aggregation operator, which is an epsilon-multilayer perceptron (E-MLP) function. E-MLP is a unique variation of the traditional multi-layer perceptron (MLP) that incorporates a learnable epsilon ($\epsilon$) parameter. This function accepts the node features and the features of its neighbouring nodes as input and adjusts the aggregation of these features using the learnable $\epsilon$ parameter in the MLP process. The E-MLP enables the model to learn the optimal balance between a node's own features and the features of its neighbours for improved graph representation learning. This parameter controls the level of expressiveness of the MLP, allowing GIN to learn more complex features. The E-MLP adjusts the contribution of each feature based on the learned $\epsilon$ value. This enables the model to adaptively decide the importance of a node's own features relative to its neighbours' features, leading to better embeddings for graph classification.

$$
h_{v}^{(k)}=\operatorname{MLP}^{(k)}\left(\left(1+\epsilon^{(k)}\right) \cdot h_{v}^{(k-1)}+\sum_{u \in \mathcal{N}(v)} h_{u}^{(k-1)}\right) .
$$

The output of MLP can be further used for various tasks like node classification, link prediction, or graph-level classification.

\subsection{Ensemble Graph Neural Networks}
Various trade-offs exist between network architecture and model performance in graph-based methods. For instance, GCNN performance may decline with added layers due to the over-smoothing of node representations, causing indistinguishable inter-class nodes. To address this issue, the jumping knowledge network is a proposed solution that selects features from more representative nodes.

GATs apply attention to all nodes, primarily focusing on immediate neighbours rather than the entire graph structure, which can result in suboptimal performance for tasks needing comprehensive graph understanding.

GINs have aggregation power equivalent to the Weisfeiler-Lehman test for distinguishing graphs. However, GIN embeddings may vary in quality for non-isomorphic graphs with similar structures and limited features.

Ensemble methods offer a solution to the individual limitations of various GNNs by utilizing a group of GNNs. Each GNN in the ensemble focuses on capturing distinct aspects of the graph's structure and information. By combining the outputs of these GNNs, a more comprehensive feature representation of the graph can be generated, which captures a wider range of information.

Ensemble methods can be a useful technique for combining the predictions of multiple GNNs to improve classification performance \cite{gens}. One common way to ensemble GNNs for classification is to concatenate the output of each GNN and use the concatenated output as the input to a final classifier. We have ensembled GCNN, GAT and GIN to get new graph embedding for classification.


We aim to leverage the flexibility of GNNs, which can be trained on graph data with varying numbers of nodes, allowing them to make accurate predictions on graphs with different node counts. This adaptability is particularly advantageous when dealing with graphs of arbitrary sizes generated through diverse methods, catering to a wide range of use cases.



In the following section, we will provide a detailed explanation of the experiments and results obtained from our analysis of the pneumonia image dataset. In addition, we will also conduct an ablation study to examine the performance of the different models used in this research.

\section{Experiments}
In this section, we introduce the dataset that was used for training the models. We then describe the various experiments that were conducted for training different GNNs.  Hyperparameter tuning was conducted using Optuna, resulting in a learning rate of 0.001, weight decay of 0.001, and a dropout rate of 0.5\% 

\subsection{Datasets}
The dataset used in our study \cite{data} is organized into three distinct subsets: training, testing, and validation, each containing subfolders for the two image categories: Pneumonia and Normal. The dataset comprises a total of 5,856 X-Ray images in JPEG format, with 1,583 images in the Normal category and 4,273 images in the Pneumonia category. Out of the 5,856 images, 1,341 Normal images and 3,875 Pneumonia images are used for training and 234 Normal and 390 Pneumonia images were used for testing.
Images were converted into graph representations, maintaining the original folder structure, and saved in pickle format for GNN training. Each experiment generated a different graph dataset.

\subsection{Models}

\begin{itemize}
\item GCN: a GCN architecture with 4 graph convolution layers was trained to utilize graph representations\ref{sec:gc} generated from images as input. Node feature extraction was performed using ResNet18, DenseNet121, and EfficientNet-B0 models. 
\item GAT: we trained a GAT model with a multi-headed attention mechanism. The architecture consisted of two layers of GAT convolution, with a dropout rate of 0.3\% applied after each convolutional layer. The learning rate was set to 0.001. 8 attention heads were utilized in each layer, allowing the model to focus on different aspects of the input graph. 
\item GIN: we trained a GIN model with 3 GIN convolution layers to generate graph embeddings. The GIN convolution layers were designed to capture the underlying structure and relationships of the graph. Additionally, a global add pooling operation was used to aggregate neighborhood features and extract higher-level representations. The learning rate was set to 0.001, and a dropout rate of 0.5\% was applied as a regularization technique to prevent overfitting. The resulting embeddings were concatenated and used for performing graph-level predictions. 
\item Ensemble Model: we combined the best performing pre-trained GNNs from the above 3 models and ensemble their results to make the classification. The learning rate was set to 0.001 with a dropout rate of 0.5\%.
\end{itemize}
The models were trained using graph datasets generated with superpixel values of 5, 10, 50, 100, 150, and 300. The features were obtained by utilizing Resnet18, EfficientNet-B0, and DenseNet121, which generated features with sizes of 512, 1280, and 1024, respectively. As a result, we have established 18 benchmark datasets for model training and classification.

\subsection{Classification}
Our primary goal was to accurately predict labels. To evaluate the performance of various models, we utilized the accuracy metric. The performance of various models on different graph datasets can be seen in Table \ref{tab:table_1}, where we present the accuracies for each conducted experiment. Table \ref{tab:table_2} shows the performance of Ensemble models. We also analyzed the model sensitivity performance across the datasets, which were generated using varying numbers of superpixels and feature extraction methods.

We achieved an optimal performance of 93.108\% in classifying pneumonia, and using DNNs, we achieved an accuracy of 92.81\% \ref{tab:my_table23}
. Our model parameters are 100 times fewer than state-of-the-art dense neural network models as shown in Table \ref{tab:table_5}, and training and inference are also faster in comparison. The 3 DNNs had an average train time of 155.702 seconds, while the 3 GNNs had an average train time of 42.759. Our results show that the model performance reached saturation after 20 epochs, and we recorded a higher sensitivity score with our pipeline compared to DNN models. 




\bigbreak



\begin{table*}
{\caption{Accuracies of Graph Neural Network Models for each Experiment}\label{tab:table_1}}%
\begin{center}
\renewcommand{\arraystretch}{1.5}
\hspace*{-1.5cm}
\begin{tabular}{ccccccccccc}
\hline
  Superpixels&  & GCNN &  &  & GAT & & &GIN &   \\
  \toprule
   & ResNet & EfficientNet & DenseNet \vline & ResNet & EfficientNet & DenseNet \vline & ResNet & EfficientNet & DenseNet & \\ 
   \hline
  5 & 0.911 & 0.883 & 0.900 & 0.892  & 0.892 & 0.897 &  0.910 & 0.900 & 0.895\\
  10 & \textbf{0.927} & 0.900 & 0.900 & 0.900 & 0.907 & 0.916 &  \textbf{0.931} & 0.921 & 0.905\\
  50 & 0.924 & 0.908 & 0.892 & \textbf{0.929}  & 0.897 & 0.875 &  0.900 & 0.911 & 0.905\\
  100 & 0.899 & 0.886 & 0.883 & 0.833 & 0.860 & 0.841 &  0.892 & 0.889 & 0.891\\
  150 & 0.878 & 0.878 & 0.870 & 0.830 & 0.842 & 0.833 &  0.895 & 0.897 & 0.908\\
  300  & 0.863 & 0.865 & 0.818 &  0.834 & 0.854 & 0.746 & 0.887 & 0.891 & 0.878 \\
  \bottomrule
\end{tabular}
\end{center}
\end{table*}

\begin{table*} 
  {\caption{Ensemble result obtained from the graph dataset created with 10 superpixels}\label{tab:table_2}}%
  \begin{center}
  \renewcommand{\arraystretch}{1.5}
  {\begin{tabular}{cccc}
  \hline
   Feature Extraction method &  Accuracy & AUC & Sensitivity \\
\hline
  DenseNet121 & 0.899 & 0.872 & 0.979 \\
  EffiecientNet-b0 & 0.852 & 0.810 & 0.992\\
  ResNet18 & 0.895 & 0.865 & 0.987 \\
  \hline
  \end{tabular}}
  \end{center}
\end{table*}



\section{Analysis}
In this section, we evaluate our graph dataset preparation method by examining alternative design options. We present the outcomes of alterations in different stages of the experiments, such as adding layers to GNN models, evaluating the quality and time consumption of two superpixel generation techniques, and comparing the feature extraction quality across three DNNs.
 
\subsection{Effects of superpixel segment number on Model Performance}

Our findings indicate that smaller image segments (a higher number of them) limit relational features, mainly due to the diffuse nature of pneumonia features in the dataset. As a result, DNNs\ref{sec:sfe} may miss out on fully capturing the image's context, adversely affecting classification accuracy, as reflected in Table \ref{tab:table_1}. Conversely, when segments are larger but fewer, their interconnectedness is better captured. This enables the CNN to capture the full context and relationships between the objects in the image, thereby enhancing classification performance.

\subsection{Effects of Superpixel methods on Model Performance}
The impact of selecting a method for superpixel creation on the performance of a GNN model is significant. All reported performance is based on the DISF method. Another significant contrast observed between the two methods was that SLIC had a shorter image segmentation time compared to DISF. This observation can be explained by the fact that DISF demands a higher initial seed value to construct superpixels. The time taken for segmentation is a crucial factor as it is one of the bottlenecks in the process of converting an image to a graph. Table \ref{tab:my_table22} contains a comparison of the time taken by each algorithm.

\begin{table}
  {\caption{Time taken (in seconds) by SLIC and DISF for segmentation}\label{tab:my_table22}}
  
  \begin{center}
  \renewcommand{\arraystretch}{1.5}%
  {\begin{tabular}{ccc}
  \hline
 Superpixels &  SLIC  &  DISF\\
\hline
5 &0.0214 & 0.1043 \\
10 & 0.0226 & 0.1007 \\
50 & 0.0298 & 0.2280 \\
100 & 0.0320 & 0.1975 \\
150 & 0.0353 & 0.3717 \\
300 & 0.0360 & 0.4833 \\
  \hline
  \end{tabular}}
  
  \end{center}
\end{table}

\subsection{Effects of Feature extraction methods on Model Performance}

DNN models like ResNet18, EfficientNet-b0, and DenseNet extract features of sizes 512, 1280, and 1024 respectively. Larger feature sizes increase sensitivity but demand more computation. While performance often correlates with feature size, this isn't consistent with graphs having edge weights. This suggests analyzing images as graphs can help models learn interconnected feature relationships.

Fine-tuning DNNs (ResNet18, Densenet121, EfficientNet-B0) for X-ray images improves GNN model performance. Table \ref{tab:my_table23} shows the DNN model's performance. Using the top model from training as a feature extractor has resulted in an average improvement of 2-3\% for all GNN models compared to using an untrained pre-trained DNN model on the pneumonia dataset.

\begin{table}
  {\caption{DNN performance on Pneumonia dataset}\label{tab:my_table23}}
  
  \begin{center}
  \renewcommand{\arraystretch}{1.5}%
  {\begin{tabular}{ccc}
  \hline
   DNN &  Accuracy  &  Parameters\\
\hline
  densenet121 & 0.9281 & 7,978,856  \\
  effiecientnet-b0 & 0.8187 & 5,288,548 \\
  renset18 & 0.9125 & 11,689,512  \\
  \hline
  \end{tabular}}
  
  \end{center}
\end{table}

\subsection{Effects of model complexity on Model Performance}
For GCNN, performance plateaus after 4 convolutional layers, and adding more causes instability due to over-smoothing, resulting in poor graph embeddings.

For GAT, while performance doesn't notably improve beyond 8 attention heads, it does increase model complexity, leading to longer training and inference times.

For GIN, performance doesn't notably improve beyond 4 convolutional layers. The choice of graph-level pooling significantly influences performance, with global add pooling outperforming mean, minimum, and maximum aggregation methods.



\begin{table}
  {\caption{Trainable Parameters in GNNs for largest feature size 1280}\label{tab:table_5}}
  
  \begin{center}
  \renewcommand{\arraystretch}{1.5}%
  {\begin{tabular}{cc}
  \hline
   GNN & Parameters\\
\hline
  GCNN & 699426\\
  GAT & 67938\\
  GIN & 99266\\
  \hline
  \end{tabular}}
  
  \end{center}
\end{table}

\section{Conclusion}
In this study, we explored various techniques for converting medical images into graphs for classification purposes. We evaluated several GNNs and compared their performance with DNNs. The following steps outline our process:
\begin{enumerate}
\item Evaluating SLIC and DISF methodologies for image segmentation.
\item Constructing a graph from the image, with each superpixel segment representing a graph node.
\item Employing various DNNs to extract features and assign them as node attributes.
\item Training and testing three distinct GNN architectures.
\item Combining the top-performing GNNs from different architectures through ensembling.
\end{enumerate}
Our results showed that GNNs performed well with 100 times fewer parameters than DNNs (refer to Table \ref{tab:table_5} and Table \ref{tab:my_table23} for comparison).

GNN models' independence from graph size enabled us to ensemble various models trained on different graph sizes, thereby improving sensitivity scores, a vital aspect in the medical field. The novelty of our work lies in the use of IFT to group like pixels together, creating nodes in a graph format. This method effectively spots spread-out irregularities, such as pneumonia, in the chest X-ray images.

In our upcoming work, we aim to test our process on a wide range of medical datasets. Given that our method achieved a 99.13\% accuracy on the MNIST dataset, we believe it has the potential for success on other datasets as well.

\section*{Acknowledgment}
This publication has emanated from research supported in part by a grant from Science Foundation Ireland under Grant number 18/CRT/6049. For the purpose of Open Access, the author has applied a CC BY public copyright license to any Author Accepted Manuscript version arising from this submission.






\vspace{12pt}

\end{document}